\documentclass{article}

\PassOptionsToPackage{numbers}{natbib}
\usepackage[preprint]{neurips_2024}

\usepackage[utf8]{inputenc} % allow utf-8 input
\usepackage[T1]{fontenc}    % use 8-bit T1 fonts
\usepackage{hyperref}       % hyperlinks
\usepackage{url}            % simple URL typesetting
\usepackage{booktabs}       % professional-quality tables
\usepackage{amsfonts}       % blackboard math symbols
\usepackage{nicefrac}       % compact symbols for 1/2, etc.
\usepackage{microtype}      % microtypography
\usepackage{xcolor}         % colors
\usepackage{graphicx}
\usepackage{float}

\usepackage[numbers]{natbib}

\title{Video Prediction Models as General Visual Encoders}

\author{%
  James Maier \\
  Carnegie Mellon University\\
  \texttt{jamesmai@andrew.cmu.edu} \\
  \And
  Nishanth Mohankumar \\
  Carnegie Mellon University\\
  \texttt{nmohanku@andrew.cmu.edu} \\
}

\begin{document}

\maketitle

\begin{abstract}
 This study explores the potential of open-source video conditional generation models as encoders for downstream tasks, focusing on instance segmentation using the BAIR Robot Pushing Dataset. The researchers propose using video prediction models as general visual encoders, leveraging their ability to capture critical spatial and temporal information which is essential for tasks such as instance segmentation.
 Inspired by human vision studies, particularly Gestalt's principle of common fate, the approach aims to develop a latent space representative of motion from images to effectively discern foreground from background information. The researchers utilize a 3D Vector-Quantized Variational Autoencoder (3D VQ-VAE) video generative encoder model conditioned on an input frame, coupled with downstream segmentation tasks.
 Experiments involve adapting pre-trained video generative models, analyzing their latent spaces, and training custom decoders for foreground-background segmentation. The findings demonstrate promising results in leveraging generative pretext learning for downstream tasks, working towards enhanced scene analysis and segmentation in computer vision applications.
  
\end{abstract}

\section{Introduction}

Over the past five years, large-scale pretraining on unlabeled data has significantly transformed the landscape of language modeling \citep{4_Naveen} \citep{5_zhang}. While strides have been made in the realm of visual understanding through models like CLIP \citep{CLIP} and BLIP \citep{blip}, which leverage multi-modal data such as organic image captions, these approaches still contend with issues related to labeler intent and bias, inherent ambiguity in text-image pairs as well as the expensive nature of labelling \citep{issues_with_DL}. However, the realm of video generation presents a distinct advantage due to the absence of ambiguity in pixel values within real video data. This suggests that video generation could serve as a well-structured task for generative pretraining of visual modeling systems, implicitly encapsulating information relevant to various downstream vision tasks such as instance segmentation, mask classification, and depth perception.

Moreover, the abundance of video data available online, estimated to constitute a large majority of internet traffic, offers a vast resource for training such models. This stands in stark contrast to text data, where even the largest datasets are dwarfed by the sheer volume of video content created daily. Given this landscape, our project aims to explore the potential of open-source video generation models as adaptable encoders for downstream tasks like instance segmentation, particularly focusing on the BAIR Robot Pushing Dataset.  Instance segmentation\citep{segmentation}, a cornerstone in computer vision applications, holds immense significance across diverse domains, from autonomous vehicles to medical image analysis. This task to classify specific objects within an image enriches systems' abilities to comprehend visual complexities and extract implicit information.

\begin{figure}[H]
  \centering
  %\fbox{\rule[-.5cm]{0cm}{4cm} \rule[-.5cm]{4cm}{0cm}}
  \includegraphics[width = 0.5\linewidth]{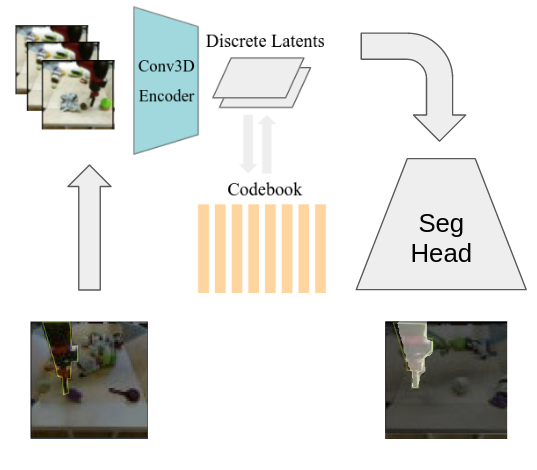}
  \caption{System Overview: leveraging pre-trained video prediction model as an encoder for downstream task of mask segmentation}
  \label{fig:intro}
\end{figure}

The primary contributions of this work are as follows:
\begin{itemize}
    \item The VideoGPT open source video prediction model is analyzed and two distinct latent spaces are proposed as inputs to downstream task-specific models
    \vspace{-.25em}
    \item The BAIR Robot Pushing dataset is selected as a testbed, and a custom dataset of 250 frame/segmentation mask pairs is created.  
    \vspace{-.25em}
    \item A variety of segmentation head neural architectures are proposed and evaluated 
    \vspace{-.25em}
    \item Results from full system compared with UNet \cite{UNET} architecture as baseline
    \vspace{-.25em}
%    \item Open source all code and custom labeled dataset (dataset: https://universe.roboflow.com/vlr-n05rz/bair ; code: https://github.com/JayMaier/VPT)
\end{itemize}

\subsection{Core Idea}
Image representation learning models are recognized for effectively capturing information within an image, including details about the scene and objects present. However, they are not equipped to represent dynamic information such as optical flow, which requires temporal context. Conversely, video representation models, exemplified by 3D inflated Convolutional layers \cite{Yamashita2018ConvolutionalNN}, excel at representing both spatial and temporal data as they can capture images along a third axis, time in our case. Specifically, they can capture critical information associated with how a scene is expected to change in the future, which is beneficial for tasks that need to understand how objects move, such as instance segmentation.

Our proposed approach aims at leveraging a latent space conditioned on a single input frame to effectively discern foreground from background information. However, a key challenge emerges: to incorporate temporal data from a standalone frame. To address this, we propose employing a 3D Vector-Quantized Variational Autoencoder(3D VQ-VAE) video generative encoder model conditioned on an input frame. We then utilize the downsampled latent space generated from the 3D VQ-VAE and feed it into a segmentation head to achieve the task of foreground vs background segmentation.

\subsection{Intuition}

At a fundamental level, we are inspired by human vision studies, particularly the Gestalt principle of common fate \citep{1_Palmer}, which suggests that humans tend to associate pixels of an image as part of the same object if they exhibit a common motion relative to each other. This concept is observed not only in adults but also in infants and individuals who have recently gained sight through experiments \citep{8_spelke} \citep{7_ostrovsky}. Based on these findings, we hypothesize that developing a latent space representative of the motion of an object from images could effectively represent crucial aspects such as object segmentation, optical flow, depth perception, object occlusion, and other dynamic characteristics of a scene.

By leveraging this latent space, we aim to capture the underlying temporal dynamics of a visual scene that are essential for tasks like object segmentation. This approach aligns with the principles observed in human visual perception, where motion and spatial relationships play a key role in perceptual organization and object recognition. Ultimately, our goal is to harness these insights to enhance computer vision systems, enabling them to better understand and interpret visual information akin to human and animal perception.

\subsection{Relation to Prior Work}

As far as our research indicates our implementation of foreground vs background segmentation using video generational models is novel, but this idea was inspired by several advancements in the fields of Generative Artificial Intelligence and the sciences of human vision:

\paragraph{Downstream Natural Language Tasks with Large language models:}
In recent years, the rapid advancement of large language models has brought about a transformative impact on the field of natural language processing \citep{3_bommasani}. By pre-training these models on extensive, unlabeled datasets of natural language, they have acquired the capacity to comprehend, manipulate, and generate human language, thereby becoming fundamental tools for a broad spectrum of downstream NLP applications \citep{2_Khadoor}, including machine translation, chatbots, and text summarization. 

Notably, there have been instances where pre-trained language models have been fine-tuned specifically for tasks such as sentiment analysis \citep{5_zhang}. In our study, we aim to extend this paradigm into the realm of computer vision by adapting and fine-tuning a suitable video generation model for the task of foreground versus background segmentation using still images.

Drawing inspiration from the success of language models in NLP tasks, our objective is to leverage the transferability of pre-training and fine-tuning techniques to enhance the capabilities of vision models. By applying similar methodologies to visual data, we seek to equip video generation models with the ability to discern foreground objects from the background in still images.

\paragraph{Representing learning from Motion and Videos:}
The human visual system exhibits a learning process that heavily relies on observing objects in motion rather than static images, as evidenced by studies conducted by Ostrovsky et al. \citep{7_ostrovsky}. Furthermore, experiments involving infants \citep{8_spelke} and newly sighted congenitally blind individuals \citep{7_ostrovsky} demonstrate that the tendency to over-segment images diminishes once objects are perceived in motion. This insight highlights the importance of temporal information in informing instance segmentation.

A notable contribution to this understanding comes from the work of Pathak et al. \citep{6_pathak}, where they successfully segmented foreground versus background and performed other tasks by leveraging low-level motion-based grouping to learn effective visual representations from images. They introduced the concept of "pseudo ground truth" by encoding motion representations derived from video data through a motion-based unsupervised learning approach. Their method involved generating representations that encapsulate motion-related information, which was then fed into established architectures for tasks such as object detection and segmentation as Fast R-CNN and AlexNet. The resulting representations exhibited strong performance in their respective tasks. The significance of this approach lies in its ability to enhance visual representation learning by incorporating motion cues derived from video data. By leveraging temporal dynamics, models can better discern object boundaries and relationships within an image, thereby improving segmentation accuracy and overall scene understanding.

%\textbf{Learning from Video Generation:}
\paragraph{Learning from Video Generation:}
The field of next-frame prediction has seen considerable research activity, with approaches that leverage pretext learning of motion and scene dynamics. Pathak et al. explored image inpainting as a means to learn scene dynamics from images \citep{13_pathak}. Wang et al. \citep{9_wang} trained convolutional networks to distinguish between pairs of tracked patches within a single video versus pairs of patches from different videos. Misra et al. \citep{10_misra} designed a network to correctly arrange shuffled frames of a video into their original temporal sequence. Another prevalent pretext task involves predicting future frames, as demonstrated by Goroshin et al. \citep{11_goroshin} predicting pixels of upcoming frames and Walker et al. \citep{12_walker} predicting dense future trajectories.

However, a key concern when utilizing generation models for such tasks is that nearby frames in a video often exhibit visual similarity. This similarity can lead these methods to focus on learning low-level features rather than the high-level information crucial for our specific task \citep{6}.

In our study, we aim to address this challenge by implementing techniques that encourage the learning of more abstract and meaningful representations from videos. By leveraging advancements in video generation and representation learning, we seek to extract rich, high-level information related to object segmentation, optical flow, and scene dynamics. This involves designing tailored architectures and training strategies that emphasize the extraction of semantically meaningful features, ultimately enhancing the model's ability to discern complex visual relationships and improve the quality of foreground-background segmentation in still images.

\section{Methodology}
Since our ultimate goal was to adapt the latent space of a video prediction model to the task of moving object segmentation, we had some specific constraints on the video model that we would start with.  First, since we wanted to run mask inference on single images (rather than a video), we needed a video model which could generate a sequence of frames conditioned on a given initial frame.  Second, since our goal was to leverage a low-dimensional representation of our input image, we needed a model with an encoder-decoder architecture.  These constraints ultimately led us to two models: MAGVIT \citep{14_magvit} and VideoGPT \citep{15_videogpt}.  Since MAGVIT is SOTA on several modern video modeling benchmarks \citep{14_magvit}, we initially chose MAGVIT.

\begin{figure}[htbp]
    \centerline{\includegraphics[width=0.5\textwidth]{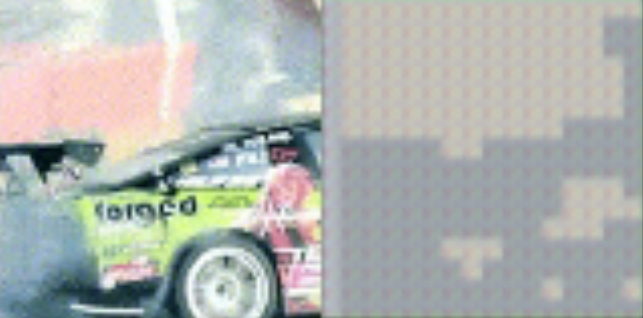}}
    \caption{Custom trained MAGVIT \cite{14_magvit} output after 10,000 steps training to fit a single frame sequence.  We were not able to get a MAGVIT model trained that would overfit to a single sequence, and switched to VideoGPT to focus on our goal of video model adaptation.}
    \label{magvit_gen}
\end{figure}

Since MAGVIT does not have publicly available weights, our effort in the project began with an effort to train a MAGVIT model from scratch that would serve as our video prediction model.  As it turned out, this effort proved to be a significant resource sink since the computational load to train the MAGVIT model was significant (In the MAGVIT paper, the authors note the model requires roughly 2 weeks to train on a V100 GPU)\citep{14_magvit}.  When we attempted to replicate the author's results we did not see good results even after copying the author's published hyperparameters and training regime(see Fig \ref{magvit_gen}), most likely due to a lack of compute (We trained the model on Geforce RTX 4070).  Since the goal of our project was to focus on adapting an existing video prediction model to downstream tasks rather than training a video prediction model from scratch, we pivoted to focus on using the VideoGPT model rather than MAGVIT.  

\begin{figure*}[]
        \centering
        \includegraphics[width = 0.9\linewidth]{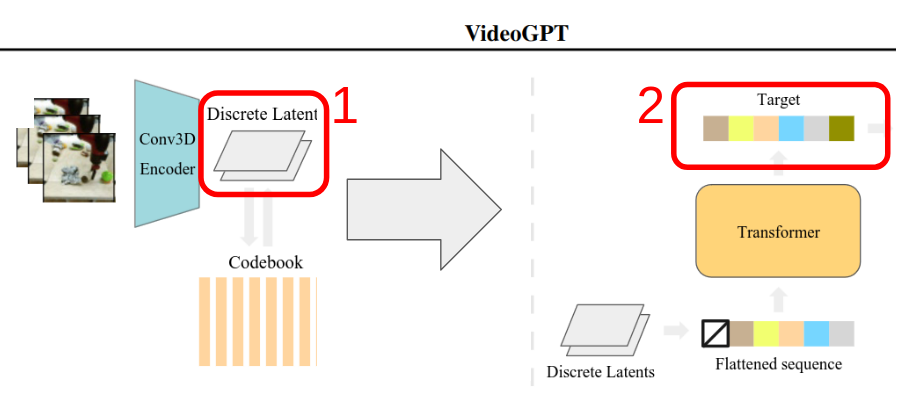}
        \caption{Two different options for latent spaces in VideoGPT to serve as inputs to our model \cite{15_videogpt}.  The box on the left labeled (1) is the output of a 3D Resnet model wihch is then fed into a transformer before being decoded.  The box on the right labeled (2) denotes the latent space defined as the output of the pretrained transformer model.  We experimented with learning segmentation masks from each of these latent spaces.}
        \label{fig:full_fled}
        \vspace{-1.1em}
\end{figure*}

Like MAGVIT, VideoGPT comprises a 3D VQVAE encoder-decoder architecture with a transformer-based sequence model operating on that latent space \cite{15_videogpt}.  Unlike MAGVIT, however, VideoGPT provides publicly available weights for a model trained to generate videos conditioned on initial frames sampled from the BAIR Robot Pushing dataset.  We conducted some initial exploratory tests on VideoGPT by pulling single frames from BAIR Robot Pushing videos and generating predicted frame sequences conditioned on these initial frames.  We saw promising results (i.e. the VideoGPT pretrained model was able to generate plausible predicted sequences from the initial frames, see [FIGURE]) and that gave us confidence that we would be able to adapt the VideoGPT model as an encoder for our downstream tasks.  Since the pretrained VideoGPT model was trained to generate video predictions from frames of the BAIR Robot Pushing dataset, we created a custom hand-labeled dataset of 250 images and segmentation masks for the BAIR Robot Pushing dataset that we would attempt to infer from the VideoGPT latent representation.

\begin{figure*}[!t]
        \centering
        \includegraphics[width = 0.9\linewidth]{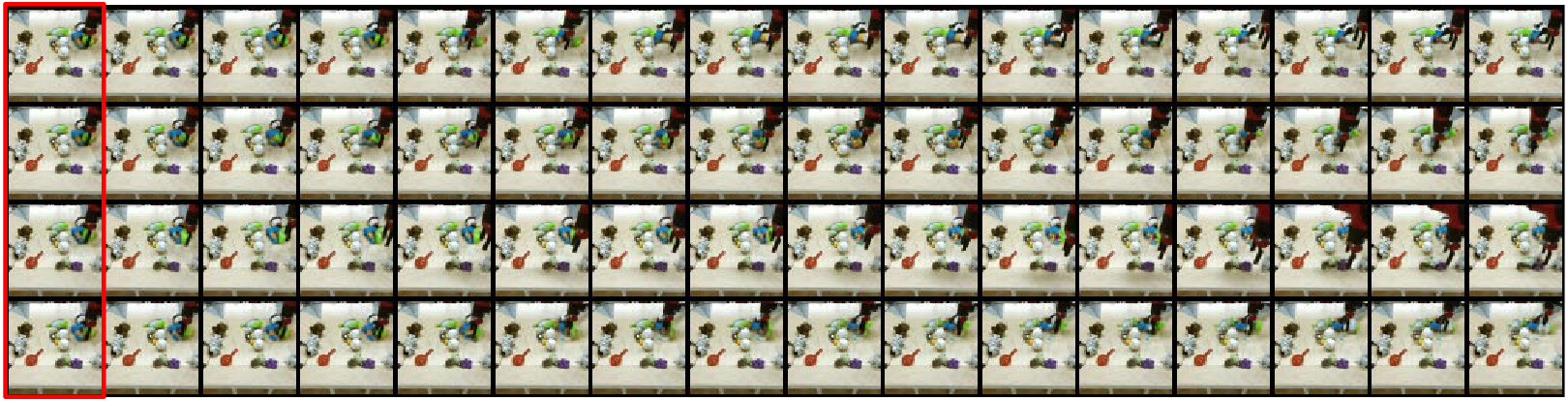}
        \caption{Bair frame sequences generated by Video GPT conditioned on same input image (left).  credit:\cite{15_videogpt}}
        \label{fig:bair_gen}
        \vspace{-1.1em}
\end{figure*}

\begin{figure}[htbp]
        \centerline{\includegraphics[width=0.4\textwidth]{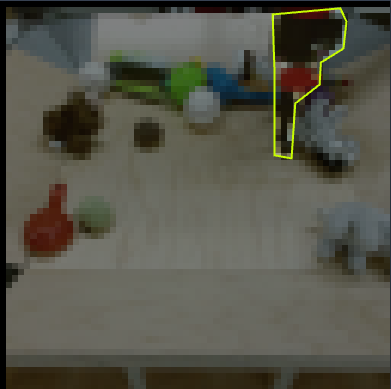}}
        \caption{Example frame from our custom dataset with robot segmentation mask}
        \label{bair_mask}
\end{figure}

In order to adapt VideoGPT as an encoder for downstream tasks, we had to decide which set of logits would provide the best latent representation.  VideoGPT, like MAGVIT, is an encoder-decoder architecture with a 3D ResNet-based encoder and decoder, and then a transformer operating on the latent space between encoding and decoding.  Because of this, we could either use the output activations of the transformer or the ResNet output activations directly as the input to our model.  Our initial intuition was that the output of the transformer would be a more processed representation of the input frame and would therefore provide superior performance, but we evaluated our system with both options.  For our segmentation network, we experimented with a number of different architectures including a linear probe, a shallow convolutional network, a fully connected layer followed by a sequence of convolutional layers, and a multi-head self attention layer followed by a sequence of convolutional layers.  We took our top performing decoder model from initial tests with a frozen encoder, and we trained a model with the same decoder architecture and a fully finetuned VideoGPT model as an encoder.  We also deployed a U-Net model as a baseline.  

In order to ensure that our model was not "memorizing the dataset", we swapped out the dataset with generated noise images paired them with segmentation masks from the real dataset.  We trained models to predict masks from noise in order to understand the representational capacity of the decoder by itself.  If we were able to achieve non-zero IOU scores with a decoder working from encoded noise, we would know that our dataset was too small/simple and we would need to label more data. The output of all of our segmentation networks was a binary mask which was intended to show the location of the robot in the input frame.

\subsection{Model Architecture and Training Details}

In the course of developing our custom decoder module we experimented with a number of different decoder architectures.  The details of these model architectures are as follows.

\begin{enumerate}
    \item Frozen 3D ResNET encoder with transfomer (pretrained from VideoGPT)
    \begin{itemize}
        \item Linear probe: input dim=4x32x32, output dim=64x64, sigmoid activation
        \item Lightweight Convolutional Decoder: Fully connected layer with output dim=4x32x32; ReLU; 2D Convolutional layer with output dim=1x32x32 and 3x3 kernel; ReLU; 2D Transpose Convolutional Layer with output dim=64x64; sigmoid activation
        \item Heavyweight Convolutional Decoder: Fully connected layer with output dim=64x64; ReLU; 4 layer convolutional encoder with output dim=4x4x512; ReLU; 4 layer convolutional decoder with output dim=64x64 and skip connections between corresponding feature maps in encoder; sigmoid activation
    \end{itemize}
    \item Frozen 3D ResNET encoder (pretrained from VideoGPT)
    \begin{itemize}
        \item Linear probe: input dim=16x16x240, output dim=64x64, sigmoid activation
        \item Lightweight Convolutional Decoder with fully connected initial layer: Fully connected layer with output dim=16x16x240; ReLU; 2D Transpose Convolutional layer with output dim=128x32x32 and 3x3 kernel; ReLU; 2D Transpose Convolutional Layer with output dim=32x64x64 and 3x3 kernel; ReLU; 2D Convolutional Layer with output dim=64x64 and 3x3 kernel; sigmoid activation
        \item Lightweight Convolutional Decoder with multi-head attention initial layer: eight head multi-head attention layer with output dim=16x16x240; ReLU; 2D Transpose Convolutional layer with output dim=128x32x32 and 3x3 kernel; ReLU; 2D Transpose Convolutional Layer with output dim=32x64x64 and 3x3 kernel; ReLU; 2D Convolutional Layer with output dim=64x64 and 3x3 kernel; sigmoid activation
    \end{itemize}
    \item Noise samples, Frozen 3D Resnet encoder with transfomer (pretrained from VideoGPT)
    \begin{itemize}
        \item Lightweight Convolutional Decoder: Fully connected layer with output dim=4x32x32; ReLU; 2D Convolutional layer with output dim=1x32x32 and 3x3 kernel; ReLU; 2D Transpose Convolutional Layer with output dim=64x64; sigmoid activation
    \end{itemize}
    \item Finetuned 3D Resnet encoder (pretrained from VideoGPT)
    \begin{itemize}
        \item Lightweight Convolutional Decoder with multi-head attention initial layer: eight head multi-head attention layer with output dim=16x16x240; ReLU; 2D Transpose Convolutional layer with output dim=128x32x32 and 3x3 kernel; ReLU; 2D Transpose Convolutional Layer with output dim=32x64x64 and 3x3 kernel; ReLU; 2D Convolutional Layer with output dim=64x64 and 3x3 kernel; sigmoid activation
    \end{itemize}
\end{enumerate}

\section{Results}

We defined our baseline as a UNET model given the miniature size of our self-annotated dataset–250 self-segmented images from the BAIR dataset performed with an IOU score of 0.83. Our best-performing model (Unfrozen 3D inflated ResNET component of the 3D-VQVAE with a lightweight Self-Attention and Convolutional layer decoder architecture) matched that score. Both these models converged to a negligible training loss within 50 thousand epochs.

\begin{table}[H]  
  \label{Result-table}
  \centering
  \begin{tabular}{lll}
    \toprule
    \multicolumn{2}{c}{Model Encoder-Decoder}                   \\
    \cmidrule(r){1-2}
    Encoder     & Decoder     & IOU Score  \\
    \midrule
    Noise Input & Linear Layer + Conv. Layers  &  0 \\
    Transformer Included & Linear Layer & 0.17 \\
    Transformer Included & Linear Layer + Conv. Layers & 0.225 \\
    Transformer Included & Linear Layer + UNET & 0.35 \\
    3D ResNET Included(Frozen) & Linear Layer & 0.17 \\
    3D ResNET Included(Frozen) & Linear Layer + Conv. Layers & 0.38\\
    3D ResNET Included(Frozen) & Transformer Layer + Conv.Layer & 0.79 \\
    3D ResNET Included(Unfrozen) & Transformer Layer + Conv.Layers & 0.83* \\
    ----Full      &      UNET---         & 0.83(Baseline)  \\
    \bottomrule
  \end{tabular}
  \vspace{0.5em}
  \caption{Ablation Study of Model Architectures: Here Transformer Included indicates that the encoder consists of both the 3D ResNET as well as the Axial-Attention (Transformer) mechanism of the 3D-VQVAE. While, 3D ResNET Included indicates that only the 3D Inflated ResNET was used the Attention Component was not used. Frozen and Unfrozen indicate the state of the encoder parameters while training.}
\end{table}

\begin{figure}[H]
\centerline{\includegraphics[width=0.5\textwidth]{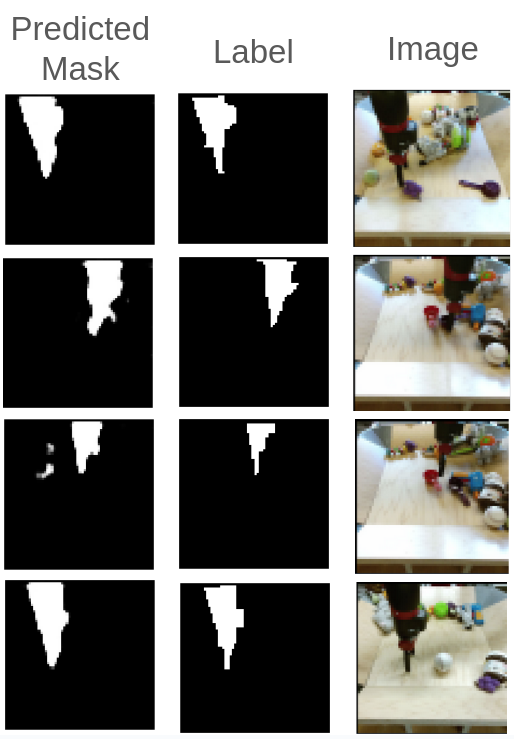}}
\caption{Example outputs from our best network (finetuned Resnet with transformer-convolutional decoder).  Left: predicted masks, Center: mask labels, Right: input images}
\label{pred_mask}
\end{figure}

\section{Discussion}
The dimensions of the latent space of the 3D VQVAE of the VideoGPT model were similar to that of a convolutional encoder– BxCxHxW– which led us to initially experiment with a simple linear layer and convolutional layers to match the output of the self annotated masks. This specific model did end up learning a segmentation mask which served as a proof of concept albeit the performance was rather poor with an IOU score of 0.225. Since our dataset was rather small we wanted to test if our 3D-VQVAE latent space was meaningful, i.e. encoded informative representation of the image. To examine this we trained the decoder on total noise to see whether our model could learn a segmentation mask, the decoder model could not learn a segmentation mask and had an IOU value of 0.0. We further attempted to add to the decoder complexity by adding residual connections, adding more convolutional layers and even attempted to develop a UNET-inspired decoder architecture. Beyond that we attempted several combinations of objective loss functions –DICE loss, Positive Weighted Binary Cross Entropy loss, and Jaccard loss– and several optimizers –SGD, ADAM, and ADAMW. Finally, we conducted a hyperparameter sweep logging our results to Weights and Biases and arrived at the best-performing model which was the UNET-inspired decoder with ADAMW optimizer. Unfortunately, upon training this model provided an IOU score of 0.35 which was far below our UNET baseline.  After this, we switched to a custom trained transformer layer instead of the pretrained one that we had been using, and we produced much better results which matched the performance of the UNET model.

\section{Future Work and Conclusion}

In this study, we demonstrate the effectiveness of utilizing generative pretext learning for downstream tasks, particularly in the context of segmentation. Initially, we harbored doubts regarding our approach of employing generative models to capture high-level scene information, given the visual similarity of adjacent frames in videos, which could potentially lead the model to learn low-level representations. However, our findings proved otherwise, as the model successfully acquired an appropriate representation of the scene.

As part of our future work, we propose the replication of this implementation using the encoder from the MAGVIT (Multimodal Attentional Generative Video Inpainting Transformer) model, which is the current state-of-the-art performance in terms of Frechet Video distance (FVD) values on several datasets \cite{14_magvit}. Leveraging such an encoder could further enhance the quality and richness of learned representations, thereby improving the effectiveness of the pretext learning approach for segmentation tasks.

Additionally, we recommend scaling up the training process by utilizing a larger and more diverse segmentation dataset, such as COCO (Common Objects in Context). Training on a larger dataset can facilitate the model's ability to generalize and capture a broader range of scene dynamics and object interactions, ultimately leading to more robust and versatile segmentation performance.

By incorporating these suggestions into future research endeavors, we aim to advance the understanding and capabilities of pretext learning methods in computer vision, paving the way for improved performance and applicability in various real-world applications of scene analysis and segmentation.

%\bibliographystyle{plainnat}
%\bibliography{bibio}

%%%%%%%%%%%%%%%%%%%%%%%%%%%%%%%%%%%%%%%%%%%%%%%%%%%%%%%%%%%%

\appendix

\section{Appendix / supplemental material}

For the link to the code as well as a partial dataset used for training please visit:
\\
\url{https://anonymous.4open.science/r/anonymous_VPT-5E85/}

\end{document}